\documentclass[11pt]{article}

\usepackage[final]{acl}

\usepackage{times}
\usepackage{latexsym}

\usepackage{amsmath}
\usepackage{booktabs}
\usepackage{threeparttable}
\usepackage{array}
\usepackage{tabularx}
\usepackage{multirow}

\usepackage{multirow}
\usepackage{graphicx}
\usepackage{placeins}

\usepackage[T1]{fontenc}

\usepackage[utf8]{inputenc}

\usepackage{microtype}

\usepackage{inconsolata}

\title{A retrospective analysis on the use of LLMs to study infant syntax learning}

\author{
	\textbf{Hélie Bazin\textsuperscript{1}},
	\textbf{Anouk Barberousse\textsuperscript{2}},
	\textbf{François Yvon\textsuperscript{3}},
	\\
	\\
	\textsuperscript{1}Sorbonne Université, Sorbonne Center for Artificial Intelligence (SCAI),
	\\
	\textsuperscript{2}Sorbonne Université, CNRS, Sciences, Norms, Democracy (SND),
	\\\textsuperscript{3}Sorbonne
	Université, CNRS, Institute of Intelligent Systems and Robotics (ISIR),
	\\
	\small{
		\textbf{Correspondence:} \href{mailto:helie.bazin_de_jessey@sorbonne-universite.fr}{helie.bazin\_de\_jessey@sorbonne-universite.fr}
	}
}

\begin{document}
	\maketitle
	\begin{abstract}
		Large language models (LLMs) have increasingly been used to investigate how children acquire syntax at an early stage of development. This is notably the central scientific goal of the BabyLM challenge, a community-wide effort to develop models that achieve human-level syntactic performance while being trained on developmentally realistic corpora. In this paper, we reflect on the use of LLMs in the study of infant syntax learning by providing an epistemological assessment of several studies from this research program. We discuss how datasets are built, which models are implemented, how they are trained and syntactically evaluated. We observe significant assumptions in the methodology of BabyLM and related studies, thus mitigating their theoretical scope. We additionally observe that using developmentally-realistic corpora have limited effects on models performance on commonly-used benchmarks, which suggest important computational differences between LLMs and the infant syntax learner.
	\end{abstract}

	\section{Introduction}\label{sec:intro}

	Syntax acquisition has been at the forefront of discussion in theoretical linguistics for several decades. How children manage to robustly acquire syntax in a few years with a limited exposure to linguistic data is still a debated issue \citep{crainNatureNurtureUniversal2001, ambridgeChildLanguageAcquisition2011}. The linguistic input of a child, called the Primary Linguistic Data (PLD), is consistent with an infinity of grammars and yet children are consistently able to select grammars with the same complex set of rules and principles. For Chomsky, this is an indication that syntax learning cannot be achieved statistically with positive evidence only, but requires a set of innate constraints on the hypothesis space of the child \citep{chomskyAspectsTheorySyntax1965}. Those innate constraints, called Universal Grammar, are language-specific biases which permit children to select the correct grammar from their impoverished PLD \citep{clarkLinguisticNativismPoverty2011}. This is the \textit{Poverty-of-the-Stimulus} (PoS) argument, one of the most debated ideas across linguistics and cognitive science.

	These debates have recently been reignited by the introduction of Large Language Models (LLMs), especially the transformer architecture \citep{vaswaniAttentionAllYou2023}. Those computational models are pre-trained on large amounts of textual data and perform syntactically and semantically at a level yet unseen in Natural Language Processing (NLP). Importantly, they seemingly do not have any language-specific bias constraining their hypothesis space. LLMs might therefore prove the possibility of learning syntax without hard constraints on the hypothesis space of the learner. This idea has been the focus of a subset of the NLP community centered around the BabyLM challenge\footnote{\url{https://babylm.github.io/}}, working at developing datasets, models and benchmarks and investigating infant syntax acquisition using small LLMs trained on developmentally realistic corpora.

	This paper questions the insights that can be drawn from this research program regarding the possibility of learning syntactic principles without domain-specific biases. In contrast to existing theoretical discussions of syntax acquisition in NLP \citep{linzenWhatCanLinguistics2018, warstadt2022artificial, wilcox2025bigger}, our work adopts a philosophy of science perspective on this research program. Although both linguistics and NLP focus on the representation and generation of syntactic structures, they ultimately investigate different target systems: the human mind and computational systems, respectively. Integrating these fields into a unified account of language acquisition therefore requires a number of assumptions and idealizations that merit careful examination. We provide an epistemological critique with the broader goal of fostering a more meaningful integration of computational approaches into the scientific study of human cognition. We discuss the construction of child-oriented corpora, the implemented architectures and the syntactic benchmarks used for studying syntax acquisition. This work is not intended to be a survey of computational studies of syntax acquisition. Rather, it is a systematic analysis of the methodology generally employed by Natural Language Processing (NLP) researchers to assess its theoretical significance. We retrospectively compare a selection of representative models across various syntactic benchmarks and draw conclusions about the use of LLMs in the study of syntax acquisition.

	Specifically, we observe that models trained on developmentally realistic corpora generally do not perform at the level of humans on syntactic benchmarks. Moreover, using developmentally realistic data has a mixed effect on LLM performance. These observations suggest important computational differences between LLMs and children that researchers should be aware of. Note that we do not engage here with the generative/usage-based debate: while our work investigates how the PoS is studied in NLP and the broader theoretical scope of this research program, we remain neutral from a theoretical linguistics perspective.

	In Section~\ref{sec:babylm}, we present the BabyLM challenge, one of the most significant effort from NLP researchers to study syntax acquisition using standardized datasets and benchmarks. In Section~\ref{sec:methodo}, we conduct a methodological survey of several PoS studies, describing how the datasets are built, the models used, and the benchmarks they are tested upon. We then compare several studies in Section~\ref{sec:results} and draw general observations on the use of LLMs to study syntax acquisition. Finally, in Section~\ref{sec:discussion}, we discuss our observations in the broader context of linguistic and cognitive modeling, questioning the effectiveness of using LLMs to study infant syntax learning.

	\begin{table*}[t]
		\centering
		\scriptsize
		\renewcommand{\arraystretch}{1}

		\begin{tabularx}{\textwidth}{
				>{\raggedright\arraybackslash}p{2.2cm}
				>{\raggedright\arraybackslash}p{3.7cm}
				>{\raggedright\arraybackslash}p{2.2cm}
				c
				>{\raggedright\arraybackslash}p{2cm}
				c
			}
			\toprule
			\centering\textbf{Paper} &
			\centering\textbf{Model} &
			\centering\textbf{Data} &
			\centering\textbf{Nb. param.} &
			\centering\textbf{TS (CDS)} &
			\textbf{Benchmark} \\
			\midrule

			\centering \citet{zhang2021you}
			&
			\centering \multirow{1}{=}{\centering MiniBERTa\\ \citep{warstadtLearningWhichFeatures2020a}}
			&
			\centering Wikipedia, Smashwords
			&
			\centering 1M, 10M, 100M
			&
			\centering 0\% (0\%)
			&
			BLiMP
			\\

			\midrule

			\centering First BabyLM
			&
			\centering ELC BERT
			&
			\multirow{2}{=}{\centering First BabyLM Corpus}
			&
			\multirow{2}{*}{\centering 10M, 100M}
			&
			\multirow{2}{=}{\centering 56\% (11\%)}
			&
			\multirow{2}{*}{BLiMP, MSGS} \\
			\centering\citep{warstadtFindingsBabyLMChallenge2023}
			& \centering\citep{charpentierNotAllLayers2023}
			& & & & \\
			\midrule

			\centering \citet{yedetoreHowPoorStimulus2023}
			& \centering LSTM, Transformer
			& \centering CHILDES
			& \centering 9.6M
			& \centering 100\% (100\%)
			& HierQ\\

			\midrule

			\centering\citet{lanLargeLanguageModels2024}
			& \centering LSTM, Transformer
			& \centering CHILDES
			& \centering 9.6M
			& \centering 100\% (100\%)
			& PG Acc., ATB Acc.\\

			\midrule

			\centering Second BabyLM
			& \centering GPT-BERT
			& \centering Second BabyLM,
			& \multirow{2}{*}{\centering 10M, 100M}
			& \multirow{2}{=}{\centering 19\% (10\%)}
			& \multirow{2}{*}{\centering BLiMP}\\
			\centering \citep{huFindingsSecondBabyLM2024}
			& \centering \citep{charpentierGPTBERTWhy2024}
			&\centering FineWeb-Edu, Cosmopedia & & & \\

			\midrule
			\multirow{4}{=}{\centering Third BabyLM \citep{charpentierFindingsThirdBabyLM2025}}
			& \centering AMLM-Hard Decay
			&  \multirow{4}{=}{\centering Second BabyLM}
			& \multirow{2}{*}{\centering 10M}
			& \multirow{2}{=}{\centering 58\% (0\%)}
			& \multirow{4}{*}{\centering BLiMP}\\
			& \centering \citep{edmanMaskYouShall2025}
			& & & &\\
			& \centering Simple Diffusion
			& & \multirow{2}{*}{\centering 100M}
			& \multirow{2}{=}{\centering 58\% (29\%)} &\\
			& \centering\citep{kosmopoulouMaskedDiffusionLanguage2025} & & & &\\

			\midrule
			\multirow{2}{=}{\centering\citet{charpentier2502babylm}}
			&\centering GPT-2
			& \multirow{2}{=}{\centering Second BabyLM}
			& \multirow{2}{*}{\centering 10M, 100M}
			&  \multirow{2}{=}{\centering 58\% (29\%) }
			& \multirow{2}{*}{\centering BLiMP} \\
			&\centering \citep{radfordLanguageModelsAre2019}& & & & \\
			\midrule

			\multirow{2}{=}{\centering\citet{padovaniChildDirectedLanguageDoes2025}}
			& \multirow{2}{=}{\centering GPT-2}
			& \centering CHILDES
			& \multirow{2}{*}{\centering 4.3M}
			& \centering 100\% (100\%)
			&  \multirow{2}{*}{\centering BLiMP, Zorro}\\
			& & \centering Wikipedia
			& &  \centering 0\% (0\%) & \\
			\midrule

			\multirow{2}{=}{\centering\citet{yangUnifiedAssessmentPoverty2026}}
			& \multirow{2}{=}{\centering GPT-2}
			& \centering Baby-F Corpus
			& \multirow{2}{*}{\centering 10M, 30M, 50M}
			& \centering 75\% (18-40\%)
			&  \multirow{2}{*}{\centering PoSH-BENCH}\\
			& & \centering Wikipedia
			& &  \centering 0\% (0\%) & \\

			\bottomrule

		\end{tabularx}

		\caption{Overview of all the papers reviewed in this study. For each paper, we specify the model, the data it is trained on and the syntactic benchmark. The models in \citet{lanLargeLanguageModels2024} are taken from \citet{yedetoreHowPoorStimulus2023}. Note that Padovani et al.\ also evaluate a RoBERTa model \citep{zhuangRobustlyOptimizedBERT2021}, but for clarity, we only feature their GPT-2 model since it performs best across tasks.}
		\label{tab:babylm-summary}

	\end{table*}

	\section{The BabyLM challenge}\label{sec:babylm}

	The BabyLM Challenge, started in 2023, is a community effort to scale down pre-training by developing high-performing language models that are trained on a developmentally realistic number of words. Researchers compete to develop architectures that perform well on specific syntactic benchmarks. The challenge typically consists of a ``Strict-Small'' track, in which models are trained on fewer than 10~million running words from a set of corpora, and a ``Strict'' track, in which models are trained on fewer than 100 million running words \citep{warstadtCallPapersBabyLM2023}. The challenge may also include additional tracks that are not purely focused on syntactic capacities. Those additional tracks, which will not be discussed in this paper, include a ``multimodal'' track, an ``interaction'' track where a smaller model learns syntax by interacting with a larger model, and more recently, a ``multilingual'' track \citep{choshenCallPapers2nd2024, charpentier2502babylm}.

	The BabyLM challenge has both engineering and scientific objectives. From an engineering perspective, it encourages the development of new, data-efficient pre-training techniques. Current LLMs are typically trained on trillions of words during pre-training and are therefore excessively expensive to train \citep{hoffmann2022training}. Techniques that improve efficiency and enable models to achieve strong performance with fewer than 100M running words during pre-training could help control training costs when scaling up to larger amounts of data. Among the scientific objectives is the goal  to advance the understanding of syntax acquisition by developing computational models that can perform syntactically at a human-level from limited training data: ``First, by reverse-engineering known and hypothetical aspects of the human learning scenario—from multimodal inputs and multi-agent interaction to innate linguistic structural biases—we can determine which factors are critical to our unique ability to learn language efficiently (...). Second, by minimizing differences between humans and models, we make results from controlled experiments carried out on models more likely to be applicable to humans" \citep{warstadtFindingsBabyLMChallenge2023}.

	\section{LLMs to study syntax acquisition: Surveying Methods}\label{sec:methodo}

	In this section, we describe the datasets, models and benchmarks employed in the study of syntax acquisition in the BabyLM challenge and related work. As mentioned in the introduction, this section should not read as fully comprehensive survey. We only include papers that we consider relevant to a broader discussion on LLMs within the fields of linguistics and cognitive science, in particular those which discuss models trained on developmentally plausible data in terms of size and content. Moreover, we only discuss the \emph{best-performing BabyLMs} in addition to the challenge baseline models, as our main interest lies in how LLMs perform on some of the most widely used syntactic benchmarks. We are aware that this excludes a number of submissions that are relevant from a cognitive science perspective, such as those discussing data ordering aspects \citep{salhanLessMorePreTraining2024, schoeneggerInfluencedrivenCurriculumLearning2025, fysikoudiActiveCurriculumLanguage2025}. Similarly, since our focus is on syntax acquisition, we leave aside various works, including BabyLM submissions, that use LLMs to study other aspects of language acquisition, such as word learning \citep{chang-bergen-2022-word}, phonology \citep{lavechinBabySLMLanguageacquisitionfriendlyBenchmark2023, goriely-etal-2024-babble} or even pragmatics \citep{askari-etal-2025-babylms}. These are valuable works and topics that should be discussed alongside this paper. Papers considered in our study are listed in Table~\ref{tab:babylm-summary}.

	\subsection{Training Data}\label{sec:data}

	\paragraph{Corpora} To replicate at best the ecological conditions children face when learning language, researchers train their models on corpora which are more or less representative of a child's PLD. A very relevant such corpus is CHILDES \citep{macwhinneyCHILDESProjectTools2000}, which contains transcripts of children-oriented conversations. Although CHILDES is often presented as a corpus of child-directed speech (CDS), a significant proportion of the data actually comprises child speech that may not be included in the PLD. Nevertheless, it is frequently used in computational studies of language acquisition as a CDS corpus \citep{huebnerBabyBERTaLearningMore2021, yadavalliSLABERTTalkPretty2023, gorielyIPACHILDESG2P2025}. For this reason, we will describe the following data as CDS, bearing in mind that this may be an oversimplification.

	Another relevant recent international initiative is BabyBabelLM,\footnote{\url{https://babylm.github.io/babybabellm/}} a multilingual collection of developmentally plausible datasets spanning over 45 languages \citep{jumeletBabyBabelLMMultilingualBenchmark2025}. Languages are regrouped into tiers depending on their data availability. Tier~1 languages have a dataset of 100M words, while Tier~2 and Tier~3 languages have datasets of 10M and 1M words, respectively. The BabyBabelLM is a significant effort to decentralize linguistics, which has traditionally focused on a few Indo-European languages.

	A subset of the BabyBabelLM including English, Dutch and Chinese forms the dataset for the new multilingual track for the fourth edition of the BabyLM challenge \citep{choshenBabyLMTurns42026}.

	However, what constitutes a developmentally plausible dataset remains unclear. This conceptual vagueness gives rise to substantial variation across corpora and studies. One important point of divergence concerns the proportion of transcribed speech (TS). The PLD is not only composed of speech instances: caretakers read books to their children, often from a very early age. Children are thus exposed to sentences that can be lexically and syntactically richer than those spoken to them \citep{montagWordsChildrenHear2015, montagDifferencesSentenceComplexity2019}. Some researchers have accordingly included children's stories\footnote{\url{https://www.kaggle.com/datasets/edenbd/children-stories-text-corpus}} or even sentences drawn from synthetic datasets of short, child-oriented narratives in their training mix \citep{eldanTinyStoriesHowSmall2023}. Yet, there is little agreement on what would constitute a developmentally realistic proportion of textual input.

	In addition, researchers also frequently incorporate non-child-oriented data. These may be textual, such as sentences extracted from Wikipedia\footnote{\url{https://dumps.wikimedia.org/simplewiki/}}, or conversational, such as movie dialogues sourced from the OpenSubtitles corpus \citep{lisonOpenSubtitles2016ExtractingLarge2016}\footnote{As a reviewer rightly observes, whether OpenSubtitles constitutes TS is debatable, since subtitles are often simply translations of the original script. Nevertheless, subtitles are conversational to a certain extent, which seems like a sufficient criterion to distinguish them from genuinely textual sources, such as books or Wikipedia.}. Once again, however, no consensus exists regarding what proportion of CDS versus other forms of input should be considered developmentally realistic. Moreover, different sources of linguistic exposure are unlikely to have equivalent effects: words heard on television may not carry the same developmental weight as those produced in parent–child interactions, where caregivers often accompany speech with gestures, shared attention, or reference to visible objects.

	This lack of agreement leads to strong discrepancies in the corpora used in PoS studies. For example, the Second BabyLM corpus contains only 58\% of TS. Researchers are furthermore allowed to use their own data, resulting in the winning submission being trained with a 1:1:1 ratio on the BabyLM corpus, FineWeb-Edu \citep{lozhkov2024fineweb-edu} and Cosmopedia \citep{allal19cosmopedia}. Consequently, its training data contains only approximately 19\% TS and 10\% CDS \citep{charpentierGPTBERTWhy2024}. Similarly, even though it is trained on 58\% of TS, the winning submission of the third BabyLM challenge for the 10M track has no CDS in its training data \citep{edmanMaskYouShall2025}. By contrast, \citet{yangUnifiedAssessmentPoverty2026} use 75\% of TS in their data, while progressively reducing the proportion of CDS as corpus size increased, such that CDS represented only 18\% of the training data once the corpus reached 50M words.. The proportion of TS even varies across the languages in BabyBabelLM: some datasets contain almost no TS, while others, notably those in Tier 3 are almost exclusively made of TS. The Dutch dataset, for instance, is used in the multilingual track of the fourth edition of the challenge despite a very small proportion of TS.

	\paragraph{Training strategies.} Most models are trained for several epochs, between 3 and 10 for the papers we review. This points to an important and yet often overlooked limitation of PoS studies. While it is true that BabyLMs and related models are trained on a limited amounts of words, a model trained during 10~epochs on 100M words, as in \citet{kosmopoulouMaskedDiffusionLanguage2025}, benefits from a training dataset not only larger in absolute size but also more structured than the infant learner who is only exposed once to the data.

	The first and second BabyLM challenges put no restriction on the number of epochs, claiming that ``from a cognitive perspective, humans have a memory of linguistic experience and can continue to access and learn from these memories.'' \cite[p.~2]{choshenCallPapers2nd2024}.

	However, the authors provide no justification for this idea, despite it introducing costly assumptions about the cognitive processes underlying syntax acquisition. The third and fourth editions of the BabyLM challenge did introduce a limitation of 10~epochs \citep{charpentier2502babylm, choshenBabyLMTurns42026}.

	\paragraph{Data ordering.} Child-oriented datasets contain data from different stages of infant development\footnote{For example, the Italian corpus in the BabyBabelLM contains textual sources that correspond to reading ages ranging from 4 to 14 years, while the educational materials include archives of high school final examinations. See \url{https://huggingface.co/datasets/BabyLM-community/babylm-ita}}. Children receive their linguistic input in a sequential order, with syntactic complexity increasing with age \cite{scarboroughIndexProductiveSyntax1990, lu2009automatic}. In order to match the learning scenario of infant learners, researchers have experimented with a training framework known as \emph{curriculum learning} \citep{bengio-etal-2009-curriculum}. In this framework, data is presented to the model in an ordered sequence of increasing complexity. Curriculum learning has been a key topic of the BabyLM challenge. Participants have experimented with various psycholinguistic metrics for data complexity, such as sentence length \citep{ghanizadeh-dousti-2024-towards}, lexical diversity \citep{mi-2023-mmi01}, and the depth of dependency trees \citep{oba-etal-2023-babylm}. However, these approaches have not led to consistent improvements in performance on BabyLM benchmarks. More recently, participants have experimented with dynamic, model-driven designs such as Influence-Driven curriculum \citep{schoeneggerInfluencedrivenCurriculumLearning2025} or Active Curriculum Language Modellng \citep{fysikoudiActiveCurriculumLanguage2025}, in which data is ordered according to its influence on the model's predictions during the initial learning phase. A key finding of the BabyLM challenge is that manually ordered data does not consistently improve performance \citep{huFindingsSecondBabyLM2024}, whereas model-driven strategies substantially impact learning \citep{charpentierFindingsThirdBabyLM2025}, even if they might lack biological or psychological plausibility \citep{schoeneggerInfluencedrivenCurriculumLearning2025}. Most papers in this strand do not put restriction on data ordering, with the exception of the work by \citet{edmanMaskYouShall2025} and \citet{kosmopoulouMaskedDiffusionLanguage2025}, who use architectures that effectively implement a sort of model-driven curriculum learning.

	\subsection{Models}

	\paragraph{Model architectures.} Most papers considered in this survey are based on transformers \citep{vaswaniAttentionAllYou2023}, though \citet{yedetoreHowPoorStimulus2023} and \citet{lanLargeLanguageModels2024} also perform their evaluation with a recurrent (LSTM) architecture. It is often preferred for language understanding tasks to use Masked Language Models (MLMs) like BERT \citep{devlin2019bert}, which can access both previous and following tokens when making a prediction, in contrast to Causal Language Models (CLMs) like GPT \citep{radford2018improving}, which perform next-token prediction based on past tokens only. \citet{zhang2021you} use the MiniBERTas from \citet{warstadtLearningWhichFeatures2020a}, a suite of RoBERTas models \citep{liu2019roberta} pre-trained on small corpora. \citet{charpentierNotAllLayers2023} use a variation of BERT, with each layer taking a combination of all previous layers as input, and not simply the sum of the input and output of the previous layer.

	However, the use of MLMs for studying language acquisition can be questioned. Not only are humans representing syntactic principles, but they are learning to generate sentences from those syntactic representations. Generating capacities should thus be expected from models of language faculty, which MLMs lack but CLMs have. Moreover, there is strong evidence that surprisal of CLMs can be effectively used to predict reading time, eye-tracking and neural data \citep{wilcoxPredictivePowerNeural2020, caucheteuxBrainsAlgorithmsPartially2022, ohComparisonStructuralParsers2022}. This suggests that the brain engages in next-word prediction when processing linguistic input \citep{goldsteinSharedComputationalPrinciples2022a}.
	To this end, we also consider studies which use CLMs. \citet{charpentier2502babylm}, \citet{padovaniChildDirectedLanguageDoes2025} and \citet{yangUnifiedAssessmentPoverty2026} use a GPT-2 architecture \citep{radfordLanguageModelsAre2019}. \citet{yedetoreHowPoorStimulus2023} and \citet{lanLargeLanguageModels2024} also use a Transformer architecture with a CLM architecture. Finally, \citet{charpentierGPTBERTWhy2024} implement a hybrid approach between CLM and MLM. Their model is essentially a MLM which, for a masked token at position $k+1$, outputs its prediction at position $k$ (like in CLMs) instead of position $k+1$. This allows to train the model with a both a causal and a masked objective, taking advantage of both approaches. GPT-BERT is a key innovation of the BabyLM challenge. \citet{edmanMaskYouShall2025} and \citet{kosmopoulouMaskedDiffusionLanguage2025} both use this architecture with an adaptive strategy for masking tokens, where easy-to-predict tokens have a lower probability to be masked.

	\paragraph{Model size.} Models in PoS studies typically have fewer parameters than standard LLMs. Pre-trained model performance has been shown to correlate with the ratio of parameters to training tokens. Smaller models thus require less parameters to reach optimal performance, as larger model might overfit the data in a  way that prevent generalization \citep{kaplan2020scaling, hoffmann2022training}. Hyperparameters for each model can be found in Appendix~\ref{sec:models}.

	\subsection{Tasks and Metrics}

	A full specification of each test benchmark with examples are in Appendix~\ref{sec:benchmarks}.

	\paragraph{Minimal Pairs.} The most common way to evaluate the syntactic capacities of LLMs is to use \emph{minimal pairs} \citep{Linzen16assessing}. Minimal pairs are pairs of sentences that differ only regarding one specific syntactic construct and are used to test the syntactic preferences of subjects. In the case of language models, researchers evaluate the syntactic ``preferences'' of a model by checking which sentence within the pair gets the higher model log-probability. The model aggregated score is then the proportion of minimal pairs for which it ``prefers'' the correct sentence.

	Whether this methodology is an indicator of genuine syntactic capacities is up-to-debate. \citet{leivadaEvaluatingLanguageAbilities2024} argue that humans can judge the grammaticality of a sentence without having to compare it to a minimally differing counterpart and are therefore hold to a much higher standard than LLMs regarding syntactic evaluation. \citet{hu2026can} put forward a theoretical assessment of this method, arguing that the log-probability assigned to sentences by LLMs is first and foremost determined by their meaning. Minimal pairs can therefore help neutralize the influence of semantic content by comparing log-probabilities of semantically equivalent sentences. We do not engage in this debate here and instead adopt the same assumption as authors who employ minimal-pair benchmarks, namely that they can be taken as indicators of syntactic capacities.

	 One of the most widely used minimal pairs benchmarks is \textbf{BLiMP}\footnote{\url{https://github.com/alexwarstadt/blimp}} \citep{warstadtBLiMPBenchmarkLinguistic2020}, which consists of 67~datasets of 1,000~minimal pairs each, testing for 12~grammatical constructs. Alternatives to BLiMP include \textbf{Zorro}\footnote{\url{https://github.com/phueb/Zorro}} \citep{huebnerBabyBERTaLearningMore2021}, which only includes words typical of the PLD, \textbf{SyntaxGym} \citep{gauthierSyntaxGymOnlinePlatform2020a} and \textbf{CoLA}\footnote{\url{https://nyu-mll.github.io/CoLA/}} \citep{warstadt2019neural}, though CoLA does not only contain minimal pairs.

	BLiMP and alternatives are not necessarily meant to study infant syntax learning. They cover core syntactic phenomenon that are likely to be found in any natural language datasets, including child-oriented corpora. But PoS arguments are typically centered around very specific syntactic phenomenon that rarely occur in the PLD \citep{pullumEmpiricalAssessmentStimulus2002, legateEmpiricalReassessmentStimulus2002, crainNatureNurtureUniversal2001, berwickPovertyStimulusRevisited2011}. It is therefore interesting to go beyond BLiMP and test the syntactic representation of models for rarer phenomenon. \textbf{PoSH-BENCH}\footnote{\url{https://github.com/xiulinyang/posh-bench}} \citep{yangUnifiedAssessmentPoverty2026} is a minimal pair benchmark which targets syntactic constructs typically mentioned when discussing syntax acquisition, like wanna-contraction or question-formation. \textbf{PG-Accuracy} and \textbf{ATB-Accuracy} are idiosyncratic names for the minimal pair evaluation performed by \citet{lanLargeLanguageModels2024} to test LLMs on complex cases of wh-question called Parasitic Gaps (PG) and Across the Board (ATB) movements.

	\paragraph{Measuring Inductive Biases.} Another way to evaluate the PoS argument using LLMs is to study their inductive biases. LLMs lack any sort of language-specific hard constraints on their hypothesis space and rather use soft, general purpose biases to generalize beyond their training corpus \citep{goldblumPositionNoFree2024, wilsonDeepLearningNot2025}. Consequently, the generalization behavior of LLMs can be compared to that of infant syntax learners to test for the necessity of language-specific biases for syntax acquisition. \citet{yedetoreHowPoorStimulus2023} thus evaluate whether LLMs align with a linear or hierarchical rule when transforming declarative sentences into yes/no questions and show a preference for the latter type of generalization (using what we will refer to as the \textbf{HierQ} benchmark).\footnote{Similar evaluations are performed by \citet{mccoy2020does, ahujaLearningSyntaxPlanting2025}, who however they do not train their model of child-oriented corpora.}

	Additionally, the Mixed Signals Generalization Set (\textbf{MSGS}) \citep{warstadtLearningWhichFeatures2020a} is a benchmark aimed to evaluate learning systems for their preference towards surface vs. syntactic generalization. A rule such as ``Does the word 'the' precede 'a'?'' illustrates the former type, while ``Is the main verb in the '-ing' form?'' illustrates the latter type.\footnote{Example from \citet{warstadtLearningWhichFeatures2020a}.} MSGS is not a syntactic benchmark per se since it does not really indicate whether the model has learned a particular construct or not. It is however useful for PoS arguments, as syntax learning might be more robust if the learner acquires early on a bias for syntactic generalization.

	\section{Retrospective analysis}\label{sec:results}

	In this section, we compare models' performance across benchmarks and draw general observations regarding the training of LLMs with developmentally plausible data. We begin with a small discussion on what should be considered a high score at minimal pairs benchmarks, before moving on to four general observations consistent across studies. A detailed specification of the scores obtained by the models at their respective benchmark is in Appendix~\ref{sec:performance}.

	Figure~\ref{fig:blimpscores} displays the scores obtained by models of varying size at BLiMP.\footnote{See also the BabyLM leaderboard: \url{https://huggingface.co/spaces/BabyLM-community/BabyLM-Leaderboard-2026}.} We use the human baseline reported by \citet{warstadtBLiMPBenchmarkLinguistic2020} as a reference, since it also constitutes the baseline for the BabyLM challenge. This baseline corresponds to the individual human agreement among 20 paid evaluators who each evaluated five pairs from each of the 67 BLiMP paradigms. The assumption that we can compare human acceptability judgments to log-probability assigned by models is controversial \citep{leivadaEvaluatingLanguageAbilities2024}. Furthermore, each evaluator only sees 335 pairs, far less than the 67000 minimal pairs that each model has to evaluate. The aggregate human score (0.96) might thus be a more accurate baseline. Still, to provide a fair assessment of PoS studies which all more or less implicitly endorse this assumption, we will consider models to have reached high performance when their minimal pair score is close to 90\%, which loosely corresponds to the human baseline in BLiMP. As one reviewer suggested, it is worth considering whether it is fair to compare the scores of models with those of adults when we are investigating infant syntax learning. However, several studies indicate that children reach adult-like syntactic processing by the age of five, having heard approximately 50 M words \citep{syntaxacqcrain, vyshedskiy2025language}. Therefore, using adult scores is not problematic and still provides a fair assessment of the models' capabilities.

	\begin{figure}
		\centering
		\includegraphics[width=1\linewidth]{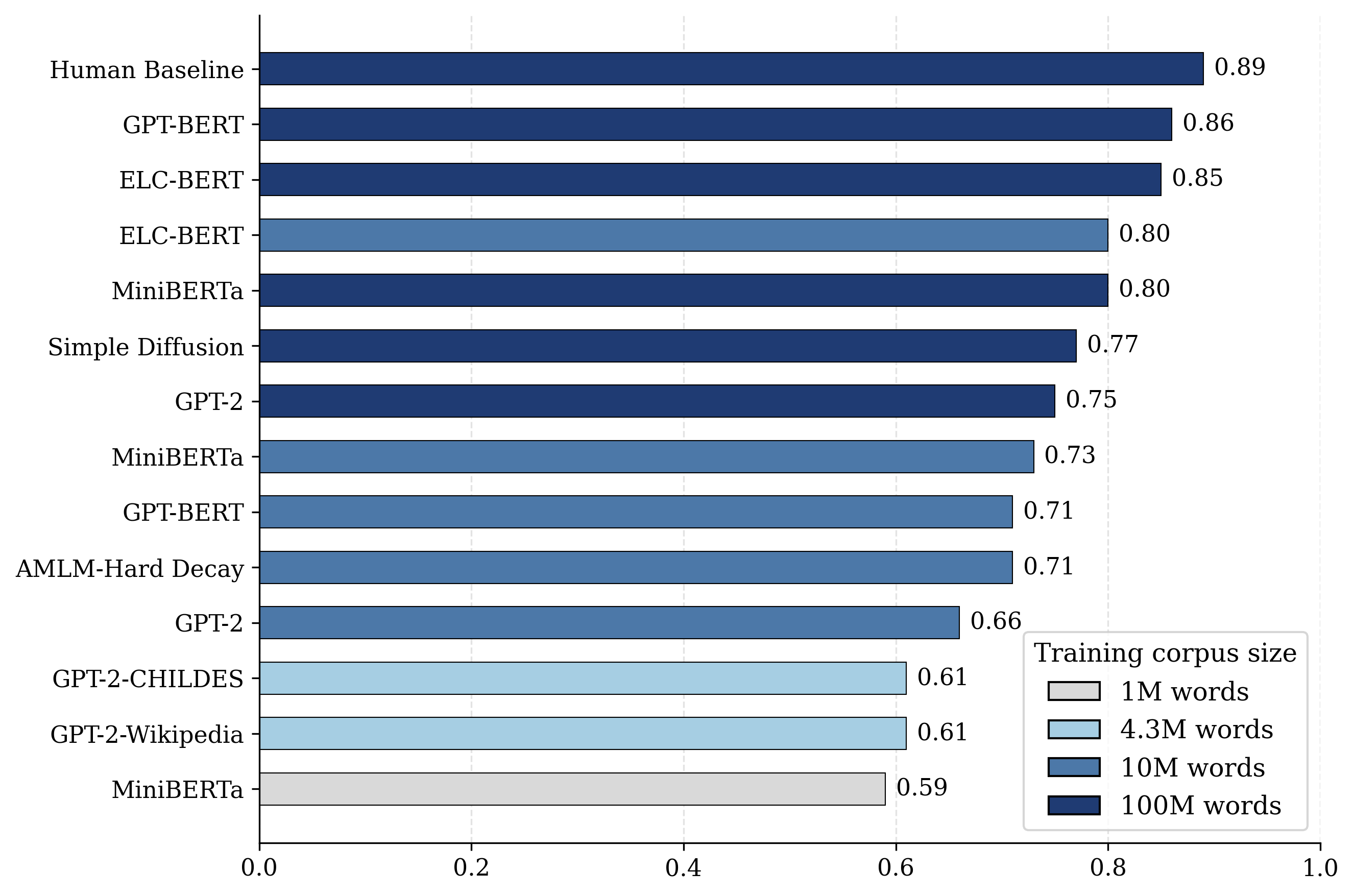}
		\caption{Overall scores obtained by the models considered in this paper when evaluated on the BLiMP, relative to the size of their training corpus, with the human baseline of \citet{warstadtBLiMPBenchmarkLinguistic2020}. MLMs perform overall better than CLMs. Performance is related to the size of the model, though this effect decreases with the corpus size. Performance gap is of the same order between models trained on less than 5M to 10M words and models trained on 10M to 100M words. See Table~\ref{tab:perfsummary} for a specification of BLiMP scores relative to the training data.}
		\label{fig:blimpscores}
	\end{figure}

	\textbf{LLMs trained on 100M words or less do not consistently perform at a human-level performance at any syntactic benchmarks.} The best scoring model on BLiMP is GPT-BERT, with a 0.81 accuracy score across minimal pairs when trained on 10M words, and a 0.86 accuracy when trained on 100M words. Accuracy drops when models are evaluated on syntactic phenomena typical of PoS arguments: GPT-2 trained on 10M words reaches a 0.65 accuracy on PoSH-Bench \citep{yangUnifiedAssessmentPoverty2026} and hardly performs above chance on question-formation tasks, regardless of the size (under 100M words) and content of the training corpus (see Table~\ref{tab:posbench} in Appendix~\ref{sec:performance}). This observation is consistent with LLMs failing to follow the hierarchical rule in \citet{yedetoreHowPoorStimulus2023}. Additionally, the models of \citet{lanLargeLanguageModels2024} can hardly perform above chance at PG and ATB movements when trained on 10M words from CHILDES (see Table~\ref{tab:transf} in Appendix~\ref{sec:performance}). For the MSGS benchmark, the winning submission of the first BabyLM displays a weak bias for syntactic over surface generalization in the Strict track; yet, the large majority of the challenge submissions, including the winning model for the Strict-small track, align with surface generalization \citep{warstadtFindingsBabyLMChallenge2023}.

	\textbf{Data scale matters, but much of the learning happens under 10M words.} This claim, found also in \citet{zhang2021you}, is consistent across studies and can be observed in Figure~\ref{fig:blimpscores}. Performance gap between models pre-trained on 10M words and 100M words is smaller than could be expected. Models trained on 100M words tend to have the same order of accuracy in areas where they already perform with 10M words. On the other hand, their performance with 100M words does not significantly improve for domains where they underperform when trained on 10M words. In \citet{yangUnifiedAssessmentPoverty2026}, models fail similarly at question-formation tasks when trained with 10M or 50M words, while their performance does not improve with data scale (and can even decrease) for wh-transformations.

	\textbf{MLMs perform better at syntactic benchmarks than CLMs.} This is clear looking at Figure~\ref{fig:blimpscores}. The best-performing models are all variations of the BERT architecture, with the winning one being GPT-BERT which benefits from both objectives. GPT-2 trained on 100M words from the BabyLM corpus is the best CLM architecture at BLiMP, reaching a score of 0.75, while ELC-BERT and MiniBERTa trained on 10M words get respectively a 0.80 and 0.73 overall score.

	\textbf{CDS (and TS) have mixed effects on performance.} This is one of the main take-aways from our retrospective study. GPT-2 models trained on 4.3 M from CHILDES or Wikipedia get the same score at BLiMP. MiniBERTas trained on Wikipedia do get a slightly higher score than GPT-2 models trained on the BabyLM corpus (see Table~\ref{tab:perfsummary} in Appendix~\ref{sec:performance}). However, this (slim) difference is most likely attributable to architectural disparity (MLMs vs. CLMs) and not to the plausibility of their training corpus. In \citet{yangUnifiedAssessmentPoverty2026}, the proportion of CDS ranges between 40\% in the 10M corpus and 18\% in the 50M, without any notable drop in performance. Performance remaining stable with the proportion of CDS decreasing could be attributed to data scaling. However, as we have seen above, increasing the corpus size does not produce significant performance gains at minimal pair evaluation. The influence of CDS is thus mixed. A similar observation can be made for TS, as models trained on 0\% TS get approximately the same score on BLiMP as models trained with a majority of TS. Conversely, GPT-2s performs better when trained on the Baby-F corpus rather than Wikipedia \citep{yangUnifiedAssessmentPoverty2026}. There is therefore no consistent pattern regarding the influence of TS and CDS across studies.

	\section{Discussion}\label{sec:discussion}

	There is no clear unified account of the role of LLMs in cognitive science. While some researchers use them as predictive models, for instance predicting brain activation from linguistic input \citep{caucheteuxBrainsAlgorithmsPartially2022}, others see them as computational cognitive models in the sense that they replicate the functional mapping of a specific cognitive system, with little regard to their biological plausibility \citep{goldsteinSharedComputationalPrinciples2022a}. In computational linguistics, two main conceptions of LLMs as scientific models can be put forward. LLMs can first be considered ``proofs-of-concepts models'', i.e., models that are used to prove the importance of specific theoretical concepts in achieving a cognitive task \citep{portelanceRolesNeuralNetworks2024}. LLMs trained on conditions similar to those met by a child when learning syntax could thus help determine whether linguistic-specific biases are necessary for syntax acquisition \citep{warstadt2022artificial}.

	Second, LLMs can be used to reverse-engineer syntax acquisition and help ``mimic infant's achievements'' \cite[p.~17]{dupouxCognitiveScienceEra2018}. Reverse-engineering, in the context of computationalism, means building systems that perform the same computations as the human brain for a specific task \citep{guestModernAlchemyNeurocognitive2025}. The goal is therefore to better understand how the brain performs the target cognitive function, by developing computational models capable of performing the same function. Under this view, LLMs are considered computational cognitive models and LLMs performing on developmentally plausible data are simply better computational model of human cognition.

	Even though the two conceptions of LLMs have distinct scientific objectives, both benefit from the BabyLM challenge and related experiments. Our retrospective study therefore yields important observations for the two.

	\subsection{Low-bias syntax learning}

	First, the claim that the PoS argument is refuted by LLMs \citep{piantadosiModernLanguageModels2024, ambridgeLargeLanguageModels2024, futrellHowLinguisticsLearned2025} does not follow from empirical NLP work, at least not from the BabyLM challenge results\footnote{We are not saying that organizers of the BabyLM challenge claim to have proven the PoS argument  wrong. The point here is addressed to researchers — notably psycholinguists and philosophers — who use the challenge as an empirical basis for such a refutation.}. When trained on smaller amounts of data, LLMs cannot consistently reach high performance on minimal-pair evaluation. They even largely underperform when the benchmark targets specific PoS phenomenon (e.g., PoSH-Bench). LLMs also systematically align with surface over hierarchical generalization. Our study shows that this observation is consistent across models, across benchmarks and across training corpora. Consequently, high consistent performance at syntactic benchmarks still requires pre-training models with inhuman amounts of data.

	LLMs are not proving the possibility of learning syntax without language-specific bias. Note that they are not proving the impossibility of learning syntax without language-specific bias or the computational necessity of Universal Grammar. The observation is simply that current models, seen as proofs-of-concept, are not proving the PoS wrong.

	\subsection{Reverse-engineering syntax acquisition}

	Reverse-engineering a cognitive system with a computational model requires that the model matches human performance at a specific task in approximately equivalent ecological conditions \citep{dupouxCognitiveScienceEra2018, warstadt2022artificial}. Not only do LLMs do not perform at a human-level at syntactic benchmarks, but it also remains unclear whether their training regime approximates the ecological conditions of infant language learning. As discussed in Section~\ref{sec:data}, there is no unified account in NLP of what constitutes a developmentally plausible corpus. We also pointed to strong simplifications in the training regime regarding number of epochs and data presentation.

	Importantly, varying the proportion of TS and CDS in the corpus does not seem to have much impact on benchmark performance. If LLMs were computational models of the infant syntax learners, reproducing—even imperfectly and with strong idealizations—the ecological conditions of syntax learning should have an impact on performance. Indeed, if a computational model replicates a cognitive specification, then it has to correlate with
	observations of the relevant human behavior \citep{guestLogicalInferenceBrains2023}. Conversely, if a model does not match human behavioral data, then it is not an accurate computational model of the relevant cognitive system. As it was rightfully pointed by a reviewer, the relevance of CDS is a debated issue in developmental linguistics. For example, \citet{cristiaChildDirectedSpeechInfrequent2019} found that Tsimane children had very limited CDS in their PLD. Conversely, there is evidence that CDS plays an important role in learning English, the primary language in NLP, whereas overheard speech has a limited impact \citep{shneidmanLanguageInputAcquisition2012, shneidmanWhatCountsEffective2013, weislederTalkingChildrenMatters2013}. Furthermore, even if the influence of CDS is more limited than previously thought, children's linguistic environment consists solely of speech. We could therefore at least expect the proportion of TS to impact performance.

	Consequently, the absence of clear performance gap between LLMs trained on CHILDES or trained on Wikipedia could indicate that CDS is as linguistically informative as adult-texts and that the inductive biases of children significantly differ from those of LLMs. Or children are sensitive to some dimension of the PLD that LLMs cannot capture, for instance whether or not the person talking to them is their caretaker. In both cases, infant learners and LLMs appear to be computationally distinct.

	\section{Conclusion}

	The BabyLM challenge is a significant step towards standardizing and augmenting NLP research by providing open-access multilingual corpora and test suites that researchers can use to evaluate their pre-training techniques. The challenge has also brought historical debates in theoretical linguistics to the forefront of computational linguistics and sparked a lively discussion among NLP researchers, linguists, and philosophers.

	However, the BabyLM challenge, and more generally the use of LLMs to study syntax acquisition, relies on important theoretical assumptions which can be questioned from linguistic, cognitive science and philosophical perspectives. In this paper, we highlight those assumptions by providing a systematic description of the training data, the training regime, the models and the benchmarks used. We showed that there was really no unified understanding of what is a developmentally realistic corpus, that the training regimes often relied on oversimplifying assumptions regarding how children learn language, that models might lack cognitive plausibility, and that the extent to which existing benchmarks measure genuine syntactic competence is still debated.

	Beyond BabyLMs and related models not yet performing at a human-level at syntactic benchmarks, we also observed no consistent pattern regarding the use of transcribed or child-directed speech. This indicates that the computational principles of LLMs differ from those of infant syntax learners, which mitigates the significance of these models in theoretical debates about language acquisition.

	Despite the negative observations we make in this paper, we do not advocate pessimism regarding the use of LLMs to study language acquisition. We fully agree with \citet{ohModelHumanLinguistic2025} that progress will most likely come from integrating NLP with empirical research on syntax acquisition architectures to build more ``human-like'' architectures. Our epistemological critique should therefore be understood as a call for deeper interdisciplinary collaboration among NLP researchers, linguists, cognitive scientists, and philosophers of science. In this respect, the BabyLM challenge constitutes an extraordinary effort—one that deserves to be continued and more fully integrated into the broader scientific study of human cognition.

	\section*{Limitations}

	We did not conduct any experiment in this paper nor did we provide solutions for the methodological limitations we identify. We also did not provide any unified measure of syntactic performance which complicates comparison between studies that use different evaluation pipelines. Furthermore, we limited ourselves to performance benchmarks. But testing for syntactic representations using structural probing or sparse auto-encoders might prove more linguistically informative. Finally, we restricted ourselves to syntax acquisition, ignoring other key aspects of language acquisition phonology, morphology, semantics and pragmatics. Those should be integrated into further study of cognitive modeling using LLMs.

	\section*{Acknowledgments} This work was supported by the National Research Agency (ANR) under the France 2030 program, with the reference number ANR-23-IACL-0007.

	\bibliography{ref}

	\appendix

	\section{Models}\label{sec:models}

	\begin{table*}[t]
		\centering
		\small

		\resizebox{\textwidth}{!}{
			\begin{tabular}{l l c c c c c}
				\toprule
				\textbf{Paper} &
				\textbf{Architecture} &
				\textbf{Parameters} &
				\textbf{Layers} &
				\textbf{Hidden Size} &
				\textbf{FF Size} &
				\textbf{Attention Heads}\\
				\midrule
				\multirow{2}{*}{\citet{zhang2021you}}& MiniBERTa MED-SMALL \citep{warstadtLearningWhichFeatures2020a} & 45 M & 6 & 512 & 2048 & 8 \\
				& MiniBERTa BASE \citep{warstadtLearningWhichFeatures2020a} & 125 M & 12 & 768 & 3072 & 12\\
				\midrule
				\citet{charpentierNotAllLayers2023} & ELC-BERT & 24 M & 12 & 384 & 1024 & 6\\
				\midrule
				\multirow{2}{*}{\citet{yedetoreHowPoorStimulus2023}}& Transformer & 43 M & 4 & 800 & - & 4\\
				& LSTM & 38 M & 2 & 800 & - & - \\
				\midrule
				\citet{charpentierGPTBERTWhy2024}& GPT-BERT & 30M & 12 & 384 & 1280 & 6\\
				\midrule
				\citet{edmanMaskYouShall2025} & GPT-BERT &30M & 12 & 384 & 1280 & 6 \\
				\midrule
				\citet{kosmopoulouMaskedDiffusionLanguage2025} & GPT-BERT &30M & 12 & 384 & 1280 & 6 \\
				\midrule
				\citet{charpentier2502babylm}& GPT-2 \citep{radfordLanguageModelsAre2019}& 124 M & 12 & 768 & 3072 & 12\\
				\midrule
				\citet{padovaniChildDirectedLanguageDoes2025} & GPT-2& 14.8 M & 8& 512 & 2048 & 8\\
				\midrule
				\multirow{2}{*}{\citet{yangUnifiedAssessmentPoverty2026}}& GPT-2-mini & ~30M & 4 & 512 & 2048 & 8  \\
				& GPT-2-small & ~110M & 12 & 768 & 3072 & 12  \\

				\bottomrule
			\end{tabular}
		}

		\caption{Hyperparameter specification of the models considered in our study. Models are smaller than industry-standard due to scaling laws. Masked Language Models and notably variation of the standard BERT architecture are the standard for natural language understanding tasks. But the recent success of Causal Language Models have motivated researchers to use variations of the GPT architecture for syntactic evaluation. We could not find the full model specification for \citet{yedetoreHowPoorStimulus2023}, which is also the only paper considered to not only use a Transformer architecture. }
		\label{model}
	\end{table*}

	Table~\ref{model} contains the hyperparameter specifications of the models we consider in our paper. Models used in PoS studies are typically orders of magnitude smaller than industry-standard, as smaller models perform better on small datasets, due to scaling laws \citep{kaplan2020scaling, hoffmann2022training}. The MiniBERTas \citep{warstadtLearningWhichFeatures2020a} are a scaled-down variation of the RoBERTa architecture \citep{liu2019roberta}\footnote{\url{https://github.com/nyu-mll/msgs}}. ELC-BERT \citep{charpentierNotAllLayers2023} is a variation of the BERT architecture found in \citet{samuel2023trained}, where instead of taking as input the previous layer, each layer takes a convex sum of all previous layers as input\footnote{\url{https://github.com/ltgoslo/elc-bert}}. GPT-BERT \citep{charpentierGPTBERTWhy2024} is a variation of the same BERT architecture, with a hybrid modeling objective\footnote{\url{https://huggingface.co/ltg/gpt-bert-babylm-base}}. When the model meets a masked token at position $k+1$, it outputs its prediction at position $k$ instead of position $k+1$ as for standard MLMs. Consequently, the output at position $k$ represents the token at position $k+1$ as in CLMs. This allows to train MLMs in a causal-fashion and benefit from both modeling method. This GPT-BERT architecture is the basis for both \citet{edmanMaskYouShall2025} and \citet{kosmopoulouMaskedDiffusionLanguage2025}\footnote{text\url{https://github.com/DespoinaKK/babylm-diffusion}}. Both study implement an adaptive strategy for masking tokens, where the probability of a token being masked is inversely proportional to its frequency. Several studies we consider use the GPT-2 architecture \citep{radfordLanguageModelsAre2019} with different number of parameters depending on the training corpus size. Finally, \citet{yedetoreHowPoorStimulus2023} develop both a Transformer model with a causal modeling objective and a LSTM\footnote{\url{https://github.com/facebookresearch/colorlessgreenRNNs}, \url{https://github.com/pytorch/examples/tree/main/word_language_model}}.

	\section{Benchmarks}\label{sec:benchmarks}

	\begin{table*}[h]
		\centering
		\small

		\resizebox{\textwidth}{!}{
			\begin{tabular}{l l c c}
				\toprule
				\textbf{Model} &
				\textbf{Corpus} &
				\textbf{Size} &
				\textbf{Score} \\
				\midrule

				\multirow{3}{*}{MiniBERTa \citep{warstadtLearningWhichFeatures2020a}}
				& \multirow{3}{*}{Wikipedia \citep{devlin2019bert}}
				& 1M & 0.59 \\
				& & 10M & 0.73 \\
				& & 100M & 0.80 \\

				\midrule

				\multirow{2}{*}{ELC-BERT \citep{charpentierNotAllLayers2023}}
				& \multirow{2}{*}{First BabyLM \citep{warstadtCallPapersBabyLM2023}}
				& 10M & 0.80 \\
				& & 100M & 0.85 \\

				\midrule

				\multirow{2}{*}{GPT-BERT \citep{charpentierGPTBERTWhy2024}}
				& \multirow{2}{*}{Second BabyLM \citep{choshenCallPapers2nd2024}}
				& 10M & 0.71 \\
				& & 100M & 0.86 \\

				\midrule

				AMLM-Hard Decay \citep{edmanMaskYouShall2025}
				& Second BabyLM
				& 10M
				& 0.71 \\

				\midrule

				Simple Diffusion \citep{kosmopoulouMaskedDiffusionLanguage2025}
				& Second BabyLM
				& 100M
				& 0.77 \\

				\midrule

				\multirow{4}{*}{GPT-2 \citep{radfordLanguageModelsAre2019}}
				& \multirow{2}{*}{Second BabyLM}
				& 10M & 0.66 \\
				& & 100M & 0.75 \\
				& CHILDES \citep{macwhinneyCHILDESProjectTools2000}
				& 4.3M & 0.61 \\
				& Wikipedia \citep{huebnerBabyBERTaLearningMore2021}
				& 4.3M & 0.61 \\

				\bottomrule
			\end{tabular}
		}

		\caption{Models performance at the BLiMP benchmark relative to their size and the training corpus. Performance is related to the size of the training corpus, but the effect of scale on performance decreases as much learning seems to happen under 10M words. Importantly, the developmental plausibility of the corpus is of limited effect on performance. Models with relatively similar architectures get similar performance regardless of the plausibility of their training corpus.}
		\label{tab:perfsummary}
	\end{table*}

	\textbf{BLiMP} \citep{warstadtBLiMPBenchmarkLinguistic2020} The benchmark consists of 67 datasets of minimal pairs. Minimal pairs are pairs of sentences that differ only regarding one specific grammatical construct. They can be used to test the syntactic preferences of subjects. For example, the minimal pair
	\begin{enumerate}
		\item \begin{enumerate}
			\item[a.] Mary likes her bike.
			\item[b.] *Mary like her bike.
		\end{enumerate}

	\end{enumerate}

	can be used to test a preference for third-person verb agreement. In the case of language models, researchers evaluate the syntactic 'preferences' of a model by checking which sentence gets the lower log-probability by the model. The 67 datasets (called paradigms) of BLiMP each contain 1000 minimal pairs. They can be organized into 12 categories regarding the syntactic construct they are used to evaluate: anaphor agreement, argument structure, binding, control, determiner-noun agreement, ellipsis, filler-gap, irregular forms, island effects, licensing, quantifiers and subject-verb agreement. The minimal pairs are generated from templates and a vocabulary of lexical items. Importantly, a minimal pair is included in BLiMP if it has reached high consensus among human subjects. A model reaching a high score on BLiMP means that its preferences match those of humans in a number of examples.

	\textbf{PoSH-BENCH} \citep{yangUnifiedAssessmentPoverty2026} The benchmark consists of minimal pairs regrouped into nine categories. The categories target four grammatical construct children acquire early on: yes/no question, island constraints, binding principles, wanna-contraction. Each categories contain 500 minimal pairs, for a total 4500 minimal pairs. The minimal pairs are also generated from templates. High score on PoS-Bench means that the model has a tendency towards the right construct across minimal pairs and categories.

	\textbf{Zorro} \citep{huebnerBabyBERTaLearningMore2021} is a syntactic benchmark very similar to BLiMP, consisting of 23 paradigms across 13 grammatical constructs: determiner-subject agreement, subject-verb agreement, anaphor agreement, argument structure binding, case,ellipsis, filler-gap, irregular, island-effects, local attractor, NPI licensing and quantifiers. Contrary to BLiMP, it only includes words that are found frequently in CDS, to avoid out-of-vocabulary items that could falsify the results.

	\textbf{PG/ATB Accuracy} This is an idiosyncratic name to refer to the evaluation performed by \citet{lanLargeLanguageModels2024} on cases of Parasitic Gaps (PG) and Across-The-Board (ATB) movement.
	\begin{enumerate}
		\item[2. ] \begin{enumerate}
			\item[a.] Which article did you file \underline{\hspace{0.5 cm}} without reading \underline{\hspace{0.5 cm}} ?
			\item[b.] *Which article did you file the report without reading \underline{\hspace{0.5 cm}} ?
		\end{enumerate}
		\item[3. ] \begin{enumerate}
			\item[a.] Which article did you file \underline{\hspace{0.5 cm}} and read \underline{\hspace{0.5 cm}} ?
			\item[b.] *Which article did you file and read the report\underline{\hspace{0.5 cm}} ?
		\end{enumerate}
	\end{enumerate}

	In 2.b., the second gap is illicit due to the adjunct clause being an island, but is made licit by the first gap in 2.a. and is therefore called a PG. Wh-movement must also occur in all conjuncts. There fore 3.b. is illicit as the movement only occurs in one conjunct. It is licit however in 3.a. This forms an example of ATB movement. PG and ATB, being complex cases of wh-movement, are the type of grammatical constructs which might appear in a PoS argument. Accuracy scores for PG and ATB cases indicate if a model has a genuine capacity at wh-movement, beyond surface regularity and 'easy' cases. PG Accuracy and ATB Accuracy consist respectively of 8064 and 6624 minimal pairs generated from a context-free grammar.

	\textbf{HierQ} \citep{mccoy2020does, yedetoreHowPoorStimulus2023, ahujaLearningSyntaxPlanting2025} refers to the evaluation of language models on question formation task. More precisely, it is used to evaluate if models align with a hierarchical or a linear rule when transforming declarative sentences into yes/no questions:

	\begin{enumerate}
		\item[4. ]\begin{enumerate}
			\item[a.] Mary who doesn't like her car does like her bike.
			\item[b.] Does Mary who doesn't like her car like her bike? (HIERARCHICAL)
			\item[c.] *Doesn't Mary who like her car does like her bike? (LINEAR)
		\end{enumerate}
	\end{enumerate}
	The model s surprisal at the first auxiliary in the transformed sentence may reveal which rule it tends to use. First, the models are trained on sentences where the hierarchical and the linear rules produce similar transformations. Next, they are evaluated on sentences where the rules generate different transformations. HierQ Score is the percentage of cases where the model accurately preferred the hierarchical rule.

	\textbf{MSGS} (Mixed Signals Generalization Set \citet{warstadtLearningWhichFeatures2020a}) is a benchmark which evaluates if a model has a bias towards surface vs. linguistic generalization. The model is trained on a corpus of labeled sentences, where labels indicate either a linguistic or a surface feature. Sentences in the training set are unambiguous, meaning that if a sentence has the label corresponding to the linguistic bias, it cannot also be consistent with the surface bias. Models then have to predict the label of ambiguous sentences. The model's label predictions can indicate if the model has a bias towards surface or linguistic generalizations.

	\section{Performance}\label{sec:performance}

	\begin{table}[t]
		\centering
		\small

		\begin{tabular}{l c c c}
			\toprule
			\textbf{Model} &
			\textbf{Corpus} &
			\textbf{Size} &
			\textbf{Score} \\
			\midrule

			\multirow{2}{*}{GPT-2 \citep{radfordLanguageModelsAre2019}}
			& CHILDES & 4.3M & 0.61 \\
			& Wikipedia & 4.3M & 0.61 \\

			\bottomrule
		\end{tabular}

		\caption{Scores for the models implemented in \citet{padovaniChildDirectedLanguageDoes2025} at the Zorro benchmark, a variation of BLiMP which only contains lexical items found in CHILDES. The developmental plausibility of the training corpus has no effect on performance.}
		\label{tab:zorro}
	\end{table}

	\begin{table}[t]
		\centering
		\small

		\begin{tabular}{l c c c}
			\toprule
			\textbf{Model} &
			\textbf{Corpus} &
			\textbf{Size} &
			\textbf{Score} \\
			\midrule

			\multirow{6}{*}{GPT-2 \citep{radfordLanguageModelsAre2019}}
			& \multirow{3}{*}{Baby-F}
			& 10M & 0.65 \\
			& & 30M & 0.64 \\
			& & 50M & 0.69 \\

			& \multirow{3}{*}{Wikipedia}
			& 10M & 0.49 \\
			& & 30M & 0.58 \\
			& & 50M & 0.57 \\

			\bottomrule
		\end{tabular}

		\caption{Scores for the models implemented in \citet{yangUnifiedAssessmentPoverty2026} at the POSH-BENCH. Though the benchmark are different, it is notable that the scores for the same GPT-2 model are lower at POSH-BENCH than at BLiMP, since POSH-BENCH is a minimal pair benchmarks which specifically targets PoS phenomena like wanna-contraction or yes/no question formation.}
		\label{tab:posbench}
	\end{table}

	\begin{table}[t]
		\centering
		\small

		\begin{tabular}{l c c c}
			\toprule
			\textbf{Model} &
			\textbf{HierQ} &
			\textbf{PG Accuracy} &
			\textbf{ATB Accuracy} \\
			\midrule

			Transformer & 0.26 & 0.31 & 0.69 \\
			LSTM & 0.15 & 0.49 & 0.21 \\

			\bottomrule
		\end{tabular}

		\caption{Scores obtained by the models trained in \citet{yedetoreHowPoorStimulus2023}. \textbf{HierQ} shows the proportion of question transformation cases where models follow the hierarchical rule rather than the linear rule. Models trained on 10M on CHILDES tend to generalize to the linear rule, indicating that their inductive biases might differ from those of humans. \textbf{PG Accuracy} and \textbf{ATB Accuracy} indicate the scores obtained by the same models on minimal pairs which target specific cases of wh-movement. The models clearly fail for those cases, which mitigates the claim that LLMs can learn syntax at a human-level under 10M words extracted from CHILDES.}
		\label{tab:transf}

	\end{table}

	The full specification of the scores obtained by various models at BLiMP can be found in Table~\ref{tab:perfsummary}. Table~\ref{tab:zorro} contains the scores obtained by models in \citet{padovaniChildDirectedLanguageDoes2025} at the Zorro benchmarks. Table~\ref{tab:posbench} contains the scores obtained by models in \citet{yangUnifiedAssessmentPoverty2026}. All scores represent overall accuracy at their respective benchmarks, i.e., the percentage of minimal pairs where the surprisal of the model was lower for the correct sentence.

	Table~\ref{tab:transf} contains the accuracy of the two models implemented in \citet{yedetoreHowPoorStimulus2023} and \citet{lanLargeLanguageModels2024}. For \textbf{HierQ}, it indicates the percentage of yes/no transformations for which the model's surprisal is lower for the auxiliary consistent with the hierarchical rule than for the auxiliary consistent with the linear rule.

	Finally, the scores at the \textbf{MSGS} benchmark use the Matthews correlation coefficient, ranging from $-1$ (systematic preference for surface generalization) to $1$ (systematic preference for linguistic generalization). A score above 0 indicates a linguistic bias. ELC-BERT trained on 10M words has a -0.01 MSGS score and therefore does not generalize linguistically. The same model trained on 10M words gets a 0.10 MSGS score, indicating a small bias towards linguistic generalization. The majority of models in the first BabyLM challenge have a negative MSGS score \citep{warstadtFindingsBabyLMChallenge2023}, which indicates that a bias for linguistic generalization over surface generalization can hardly be acquired under 100M words. The MSGS track was dropped in the following challenges.

\end{document}